%% file: main.tex
\documentclass{article}
\usepackage{ijcai22}

\usepackage{times}
\usepackage{soul}
\usepackage{url}
\usepackage[hidelinks]{hyperref}
\usepackage[utf8]{inputenc}
\usepackage[small]{caption}
\usepackage{graphicx}
\usepackage{amsmath}
\usepackage{amsthm}
\usepackage{amsfonts}
\usepackage{booktabs}
\usepackage{algorithm}
\usepackage{algorithmic}
\usepackage[dvipsnames]{xcolor}
\title{Inkorrect: Online Handwriting Spelling Correction}

\author{
}

\author{
Andrii Maksai \and
Henry Rowley\and
Jesse Berent\And
Claudiu Musat
\affiliations
Google Research 
\emails
\{amaksai, har, jberent, cmusat\}@google.com
}

\begin{document}

\newcommand{\amaksai}[1]{\textbf{amaksai: }\textcolor{red}{#1}}
\newcommand{\cmusat}[1]{\textbf{cmusat: }\textcolor{blue}{#1}}
\newcommand{\har}[1]{\textbf{har: }\textcolor{green}{#1}}
\newcommand{\jberent}[1]{\textbf{jberent: }\textcolor{pink}{#1}}

\maketitle

\begin{abstract}
We introduce Inkorrect, a data- and label-efficient approach for online handwriting (Digital Ink) spelling correction -- DISC. Unlike previous work, the proposed method does not require multiple samples from the same writer, or access to character level segmentation. 

We show that existing automatic evaluation metrics do not fully capture and are not correlated with the human perception of the quality of the spelling correction, and propose new ones that correlate with human perception. We further show that our method outperforms existing work, using new automated metrics and human evaluation. 

We additionally surface an interesting phenomenon: a trade-off between the similarity and recognizability of the spell-corrected inks. We further create a family of models corresponding to different points on the Pareto frontier between those two axes.
We show that Inkorrect’s Pareto frontier dominates the points that correspond to prior work.
\end{abstract}

\input{introduction}

\input{related}
\input{evaluation}
\input{method}

\input{results}

\section{Conclusion}

We proposed Inkorrect, the first DISC method that balances recognizability and similarity in digital ink spelling correction. It is data-efficient and privacy-enabling, as it does not need multiple samples from the same writer, and resource-efficient, removing 
dependencies on character segmenters. 

We study the tradeoff between various recognizability and similarity measures and introduce DISC-specific ones ($CDE$), that accommodates the expected differences coming from correcting the spelling. 
We show these correlate well with human judgment and use them to evaluate Inkorrect. Its  Pareto frontier dominates the prior work.

%

\bibliographystyle{named}
\bibliography{ijcai22}

\end{document}

%% file: introduction.tex
\section{Introduction}

Digital Ink offers new experiences. From writing to drawing and full-page note taking, it enables new functionalities, for instance complementing user inputs with synthetic ones. We study digital ink spelling correction (DISC), which, alongside ink beautification and autocompletion, belongs to a family of methods known as Digital Ink Synthesis.

\paragraph{Digital Ink is complex.} Writing is formed by chaining multiple strokes,  each stroke being a sequence of distinct points or curves. This complexity encodes a set of human-observable properties: the content, for instance characters, words, gestures or drawings; the writer’s style -- that makes a person’s writing unique; the prosody, that makes a certain word or symbol different from others written by the same writer. Successful ink synthesis includes all the factors above.

\begin{figure}
\label{fig:inkorrect}
\includegraphics[trim=50 20 40 20,clip,width=0.48\textwidth]{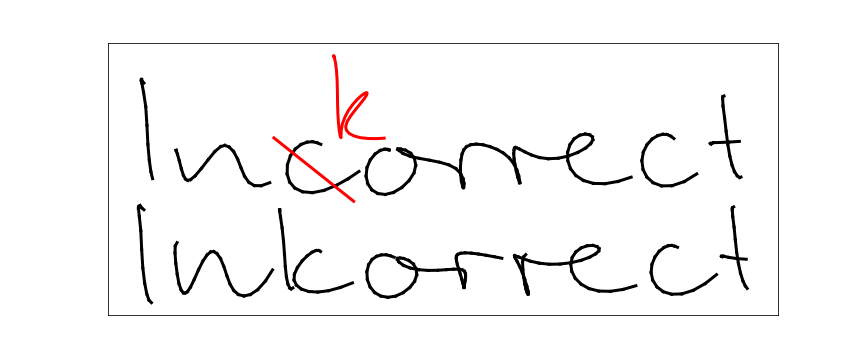}
\caption{High fidelity ink correction using our approach.}
\end{figure}

\paragraph{Ink spelling correction is valuable.}
Let’s assume the word “Mi\textbf{st}ake” is misspelled as “Mi\textbf{ts}ake”. Pen and paper error correction implies the manual chaining of three operations - finding the error, partially deleting the affected strokes and replacing them with novel ink. In DISC, these steps are automated, with all user content staying in ink form. The user does not need to switch between ink and rendered text to correct errors, providing an incentive to continuously use the writing device.

\paragraph{Human-generated ink correction is hard.}
While there is a large amount of research on handwriting synthesis, only Deepwriting (DW)~\cite{aksan2018deepwriting} explicitly mentions DISC. Several factors make spelling correction a harder task. 
Training a generative model conditioned on the original ink and the spell corrected label to generate exactly the spell corrected ink requires access to (original, spell-corrected) ink pairs. Collecting those, either with a mechanism that allows deleting and reconnecting strokes, or by asking users to rewrite their own sample, requires a complex setup, and may fail to produce pairs with high enough similarity. It is even harder if the spelling correction results in changes of position of other words in the line, requiring rewriting the whole line. Acquiring diverse data from multiple writers adds to the task difficulty and cost. Achieving comparable results without the need for cumbersome data collection is desirable.

\paragraph{Style transfer is insufficient for DISC.}
Existing methods resort to style transfer, extracting style from the original ink, and using it to generate the spell-corrected one. We argue that style transfer, while preserving the general similarities of the writing, does not fully capture the way the writer has written a particular ink. Additionally, style transfer typically needs multiple samples from the same writer, or multiple characters from the same writer via character segmentation, which makes the data collection costlier. Furthermore, character segmentation creates a dependency on an external system.

\paragraph{The Similarity -- Recognizability tradeoff.} 
While preserving the writer style is important, a key trait of DISC is that similarity is not enough. The corrected ink needs to be recognized as different from the original, namely as the spell-corrected label. Figure~\ref{fig:tradeoff} shows examples of the cases where those two goals can not be achieved simultaneously, surfacing the tradeoff between them. 

Existing similarity-based evaluation is thus inappropriate for DISC, as it misses the tradeoff between similarity and recognizability. Measuring the quality of DISC was until now (in DW) done manually via the mean opinion score, which, like all human-based evaluation, is prohibitively expensive to use with multiple model versions or hyperparameter tuning.

\paragraph{Our contributions}
\begin{itemize}
    \item We propose a way of measuring the quality of DISC via two axes (recognizability and a new metric of similarity with the original ink). We show that these correlate with human perception. We study the trade-off between similarity and recognizability. 
    \item We create a segmentation-free, data-efficient, privacy-enabling DISC method. It uses unpaired data without writer IDs and does not require character segmentation. It thus does not require an expensive data collection or annotation process. 
    \item We show that our approach is preferred to the existing work by both automatic metrics and human evaluation. It outperforms the state of the art on both recognizability and similarity. 
\end{itemize}

%% file: related.tex
\section{Related work}

There is a wide variety of methods for generating synthetic ink. A subset of those use style transfer, a widely used ingredient in spelling correction. A single method mentions DISC as an explicit application.

\paragraph{Generative models} for online handwriting data have been of interest to the scientific community, starting with the seminal work ~\cite{graves2013generating}. Handwritten text, sketches, and diagrams have been generated with various
RNNs/LSTMs/VRNNs~\cite{ha2017neural,aksan2018deepwriting,chang2021style} and Transformers~\cite{ribeiro2020sketchformer,aksan2020cose}.
~\cite{song2018learning} applied a CycleGAN-based method to map between the image and sketch domains, with the underlying handwriting generating model being an LSTM, further extended with a CNN-based autoencoder by~\cite{cao2019ai}.


\paragraph{Style transfer and Prosody.}
The writer’s style is central in synthesis, as it underpins three major applications: beautification, autocompletion and spelling correction.
~\cite{graves2013generating,aksan2018deepwriting} prime the model on a particular input, which implicitly encodes the style representation in the hidden state of an RNN. 
~\cite{chang2021style,kotani2020generating} have an explicit style extraction model which produces a vector representing of style, which can later be used as an input to the generative model.~\cite{kotani2020generating} use both global and per-character style components, which are extracted with the help of character segmentation.

The writer’s overall style is not sufficient to capture the specifics of the characters or words to be corrected. 
While this issue has received relatively little attention so far in online handwriting synthesis, it has been addressed in speech synthesis by combining the speaker identity with prosody.
The latter is defined as \textit{variation in signals that remains after accounting for variation due to phonetics, speaker identity, and channel effects}~\cite{skerry2018towards,raitio2020controllable}.

\paragraph{Spelling correction}
Style is the central element in the only work addressing DISC, DW~\cite{aksan2018deepwriting}. Here, the authors detail a conditional generative model using a VRNN~\cite{chung2015recurrent} as the backbone for synthesis.
They use a handwriting style representation for every letter, which explicitly decouples the position of the letter from the way it is written, allowing generation of spell-corrected writing in which the unchanged letters will be quite similar to how they were written in the original ink. However, this approach needs character segmentation of the input, which can be error-prone and hard to obtain. Furthermore, as authors themselves note, this approach does not perform well for samples with delayed strokes and cursive writing, likely because it assumes monotonic segmentation. 

\paragraph{Evaluation.}
As detailed in~\cite{theis2015note}, the negative log-likelihood of synthetic samples, their visual fidelity, and automated performance metrics can be largely uncorrelated from one another. 
As human evaluation of every model is prohibitively expensive, it was replaced for generative models in image, audio, video, and text domains with automated metrics such as Fr\'echet Inception Distance (FID)~\cite{heusel2017gans}, Fr\'echet Audio Distance~\cite{kilgour2019frechet}, FVD~\cite{unterthiner2018towards}, and ROUGE~\cite{lin2004rouge}. In most cases, however, a precondition is the access to an expensive dataset of human-generated labeled outputs. 

In online handwriting generation, there are 3 general approaches to evaluating the quality of the system:

\textbf{Human evaluation} to measure the quality of the output data, typically on a 5 point scale, as done by~\cite{aksan2018deepwriting,ribeiro2020sketchformer,cao2019ai,chang2021style,das2021cloud2curve}.

\textbf{Recognizability}, measured on the generated samples with a separately trained recognizer~\cite{cao2019ai,ribeiro2020sketchformer,chang2021style,das2021cloud2curve}.

\textbf{Similarity}, measured between the output of a generative model and a real sample, often to measure the fidelity of reconstruction.~\cite{song2020beziersketch,das2021cloud2curve} used a variation of Earth Mover's distance, Sliced-Wasserstein distance~\cite{kolouri2018sliced} as a loss to minimize, and FID to measure the quality of reconstruction of the sketches. ~\cite{aksan2020cose} used Chamfer distance, a measure from the field of differential geometry used to align two point clouds. In Section~\ref{sec:evaluation} we outline the shortcomings of these metrics for DISC and propose an alternative.


%% file: evaluation.tex
\section{Evaluation}
\label{sec:evaluation}

\begin{figure}
    \centering
    \includegraphics[trim=10 10 10 10,clip,width=0.48\textwidth]{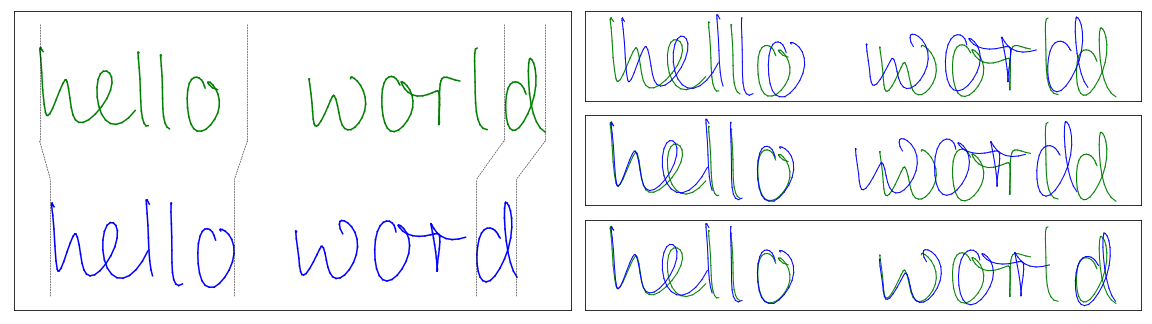}
    \caption{Comparison of $\mathit{CD}$, $\mathit{CDO}$, and $\mathit{CDE}$. \textbf{Left}: Original and spell-corrected ink. Dotted lines indicate shifts needed to get the optimal alignment of ink pieces found when optimizing $\mathit{CDE}$ with $K=3$. \textbf{Right}: Original alignment of the inks used to compute $CD$ (top), alignment found by $\mathit{CDO}$ (middle), and $\mathit{CDE}$ (bottom). Since words start at a different position, $CD=10.9$ shows that without shifts alignment is poor; $\mathit{CDE}=6.0$ aligns first word well, but that results in a poor alignment for the second one; $\mathit{CDE}=3.7$ finds a reasonable alignment for both words.}
    \label{fig:alignment}
\end{figure}


Creating high-fidelity paired samples of original and spell-corrected inks is difficult. While creating spell-corrected labels is trivial, spell-corrected inks are inaccessible. We therefore introduce a method to evaluate the quality of handwriting spelling correction 
only having access to a pair of (original ink, original label, spell-corrected label). 

The two main properties of the spell-corrected ink are: (a) it must be possible to read and recognize the spell-corrected label, and (b) it must be similar to the original writing. 
We assume that, as in DW, we evaluate a conditional generative model that takes an original ink and a spell-corrected label, and produces a spell-corrected ink.

\subsection{Recognizability}
Similar to many previous works, we use a recognition model trained on a separate dataset to evaluate how recognizable the synthetic ink is. Similar to~\cite{chang2021style}, we use Character/Word Error Rate ($\mathit{CER}$/$\mathit{WER}$) as our metrics for text recognizability.

\subsection{Similarity and Distances between Inks}
Typical distance metrics between the original and reconstructed ink are either in the input space, like Chamfer Distance($CD$)
, or in the feature space, like FID.

In DISC however, we don’t want a perfect reconstruction~-- rather, we want most of the ink to be reconstructed, except for the parts that have been spell-corrected. 
A solution is to segment the original and reconstructed ink into characters, and measure the distance only between the characters on the common subsequence of the two labels. However, this makes the metric dependent on the success of a separate segmentation model.
We propose an alternative - a relaxation of the $CD$, to better capture the distance between the original and spell-corrected ink, without any external dependencies.

\textbf{Chamfer distance}, used in ~\cite{aksan2020cose}, is a metric from the differential geometry field, which computes the distance between the two point clouds $P$ and $Q$~\cite{dougherty2018mathematical}:
$$CD(P,Q)=\frac{1}{|P|}\sum_{p\in P}\min_{q\in Q}||p-q||_2 + \frac{1}{|Q|}\sum_{q\in Q}\min_{p\in P}||p-q||_2 $$

However, if we use it when we do the spelling correction, where the positions of the letters may shift due to letters being introduced or removed, exact point-to-point correspondence is lost and the metric loses its meaning. 

We propose a modification of $CD$ that accounts for spelling correction. Below we use $P=\{p_i\in \mathbb{R}^2|1\le i \le |P|\}$ and $Q$ to identify the points of the original and spell-corrected inks, ordered by \textit{the x-axis of the points}. For inks, written nearly horizontally, in right-to-left languages, this means earlier points roughly correspond to earlier letters.

\textbf{Optimizing global offsets.} To better align inks shifted w.r.t. each other, we allow arbitrarily shifts to minimize the pairwise distance. This is similar how to it is used to align two point clouds in differential geometry, either by either ICP~\cite{arun1987least}, or by grid search of all possible offsets. This work uses a grid search, since we do not need to optimize scale or angle, but only translation.
We denote this Chamfer Distance Offset, or $\mathit{CDO}$ for short. 

$$\mathit{CDO}(P,Q)=\min_{o \in \mathbb{R}^2} CD(\{p+o|p \in P\}, Q)$$

\textbf{Optimizing local offsets.} To align ink parts that may be shifted w.r.t. each other, we allow both inks to be split into $K$ corresponding groups $(P_1,\ldots,P_k)$, $(Q_1,\ldots,Q_k)$. Each point in the ink should belong to one of the groups, each group should contain a set of points with consecutive indices (that is, all points with x-axis in a certain range), and we perform matching in each group independently:
$$\mathit{CDE}(P,Q,K)=\min_{(P_1,\ldots,P_k) ,(Q_1,\ldots,Q_k)}\sum_{i=1}^K \mathit{CDO}(P_i,Q_i)$$

Chamfer Distance Edit-Aware ($\mathit{CDE}$), has following properties: \textbf{(i)} $\mathit{CDE}(P,Q,1)=\mathit{CDO}(P,Q)$ \textbf{(ii)} $\mathit{CDE}(P,Q,K+1)\le \mathit{CDE}(P,Q,K)$. 

\paragraph{Selection of $K$} We observe that two main sources of ink miss-alignments are characters added or deleted during correction and shifts due to offsets between individual words. (See Figure~\ref{fig:alignment} for example). Therefore, we select $K$ to be exactly that: number of words in the original label plus edit distance between original and spell-corrected label.

\paragraph{Optimizing $\mathit{CDE}$} Since $CDE$ considers splits into consecutive point segments, the optimal value of $\mathit{CDE}$ can be found via dynamic programming to compute $F_{|P|,|Q|}^K$ where $F_{i,j}^k$ is the distance when grouping the first $i$ points of $P$ and first $j$ points of $Q$ into $k$ groups to minimize a sum of $\mathit{CDO}$:

$$F_{i,j}^k=\min_{l<i,m<j}F_{l,m}^{k-1}\nonumber \\+\mathit{CDO}(\{p_l,\ldots,p_i\},\{q_m,\ldots,q_j\})$$
Figure~\ref{fig:alignment} shows an example alignment of inks found by $\mathit{CDE}$.

\textbf{Limitations.} $\mathit{CDE}$ assumes inks are written nearly horizontally and are of the same scale (which is the case for datasets we used). In the general case external line height and writing angle estimation models may be needed.

\subsection{Recognizability-similarity Tradeoff}
\label{subsec:tradeoff}

\begin{figure}
    \centering
    \includegraphics[trim=14 10 10 10,clip,width=0.48\textwidth]{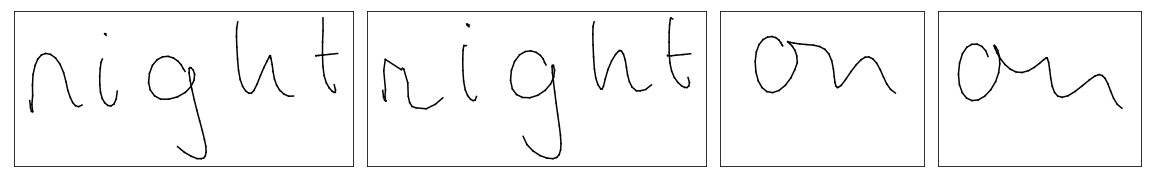}
    \caption{Similarity-recognizability tradeoff example: pairs of inks that are so similar that it’s hard to recognize the second one correctly (is the second one “right” or “night”? Is the last one “on” or “an”?). 
    }
    \label{fig:tradeoff}
\end{figure}

Recognizability and similarity are not independent for spelling correction. Perfect similarity requires having the spell-corrected ink be identical to the original - with the original misspelling not fixed. Conversely, if the original ink is hard to recognize, making the spell-corrected ink perfectly recognizable necessarily means making it different from the original. Figure~\ref{fig:tradeoff}  supports this intuition.  

\subsection{Human Evaluation Protocol}
\label{subsec:protocol}
We presented the users with a triplet of samples (original ink, spell-corrected ink from Inkorrect, spell-corrected ink from DW) and asked: \textit{“which of the two versions is a better spelling correction? If you had a system that would be able to replace the original ink with a corrected version, which one would you prefer?”}
This helps us answer three questions:
\begin{enumerate}

\item Does the recognizability metric correlate with human preference? 
\item Which similarity metric correlates best with human preference?
\item Is our DISC approach preferred to the existing baseline?
\end{enumerate}

After evaluation, users were asked to reflect on the criteria they used for selecting the best spelling correction.

We answer these questions in Section~\ref{sec:results} by analyzing the correlations between the human evaluation and the per-sample values of $\mathit{CER}$, $\mathit{CDE}$, and $\mathit{CDO}$.

%% file: method.tex
\section{Method}
\label{sec:method}

\begin{figure}[b]
    \centering
    \includegraphics[trim=50 160 20 5,clip,width=0.48\textwidth]{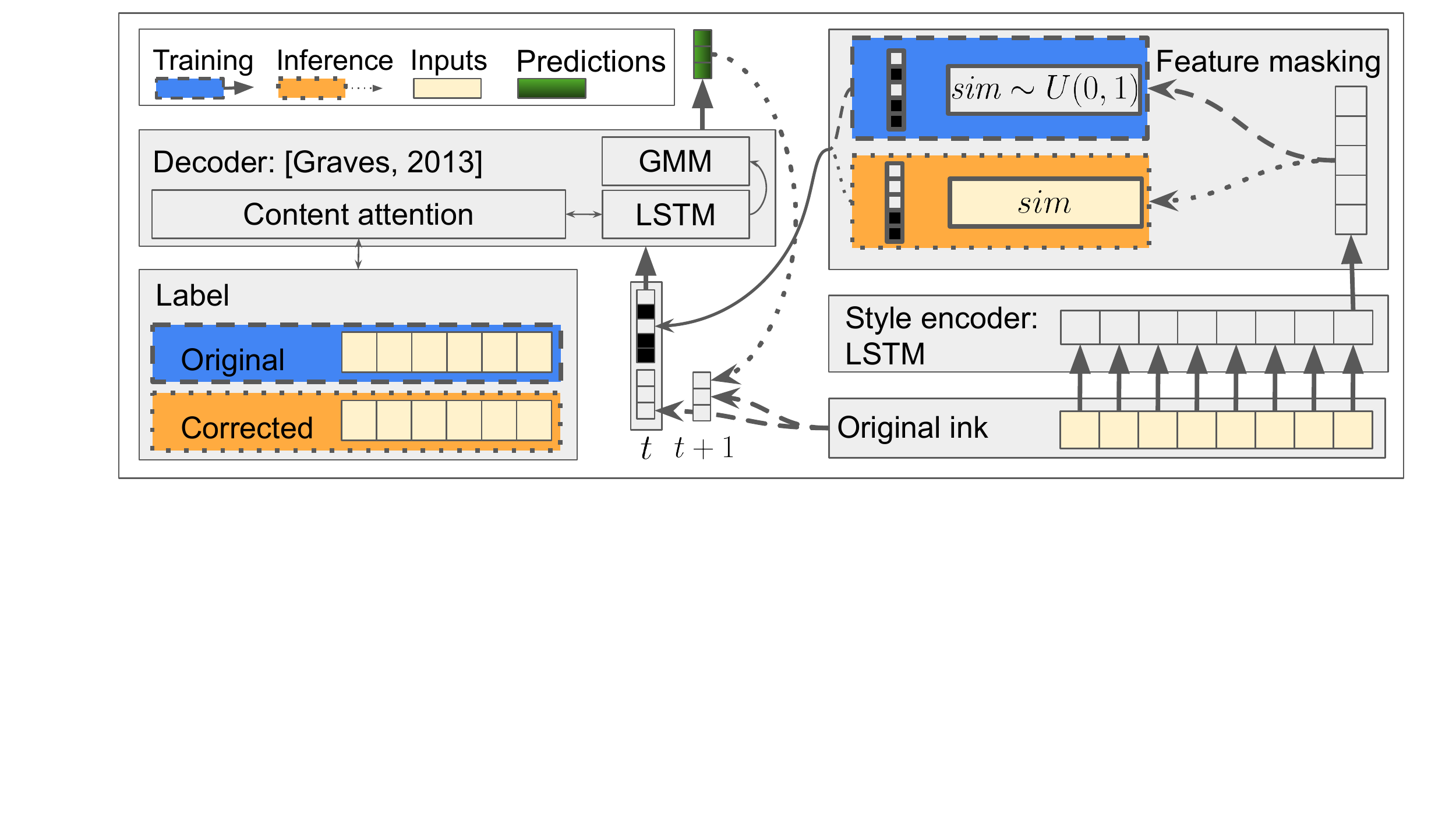}
    \caption{Proposed architecture.}
    \label{fig:architecture}
\end{figure}

Proposed model is summarized in Figure~\ref{fig:architecture}.

\textbf{Generative model} Our backbone generative model is a multi-layer LSTM with monotonic attention over the label which generates the parameters of GMM distribution for the next point in the ink~\cite{graves2013generating,chang2021style
}.

\textbf{Style extraction} Inkorrect differs from DW in the way the original ink is represented, in the form of the latent representation, and in the model that is used to extract the latent representation. A key element in our model is the way the \textbf{style} of the original ink is extracted. Empirically, style captures both global features of the original ink (size, angle, cursiveness, etc.) and local features (ways in which individual letters are written), allowing spelling correction which is very similar to the original ink, while at the same time enabling model to generalize to labels that don’t match ink content.

\begin{figure}
    \centering
    \includegraphics[trim=90 40 50 40,clip,width=0.5\textwidth]{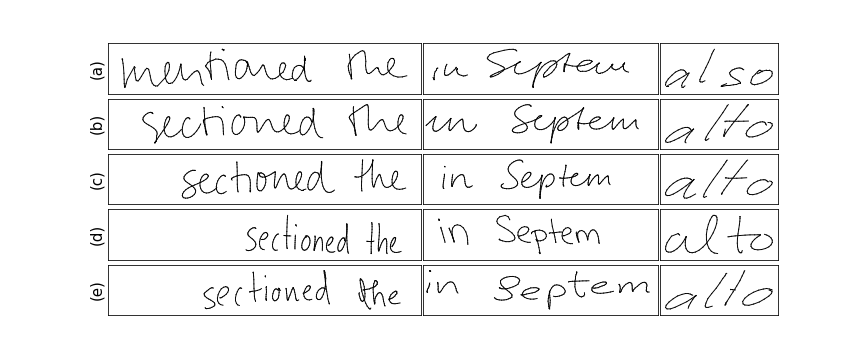}
    \caption{Effect of similarity masking and comparison with DW. \textbf{(a)} Original ink; \textbf{(b)} \textit{sim} 1.0; \textbf{(c)} \textit{sim} 0.4; \textbf{(d)} \textit{sim} 0.0. \textbf{(e)} DW. Similarity gradually decreases as we increase the masking of the style.}
    \label{fig:comparison}
\end{figure}

Our style extraction model takes a whole original ink as input and outputs a fixed dimension representation of it, which is then appended to the input to every time step. We use a two layer LSTM~\cite{hochreiter1997long} (although any other model to process the sequence input could be used, such as a Transformer~\cite{vaswani2017attention}) to compute the style. Taking only the ink itself as the input, our approach does not require knowledge of the character segmentation, nor does it assume monotonicity. A fixed dimension latent representation captures not only local character-level features, but also global ink-level features. This allows it to provide a spelling correction more similar to the original ink, as shown in Figure~\ref{fig:comparison}.

\textbf{
Tradeoff Control through Masking.}
The main novelty is a feature masking layer that can mask features of the style vector. 
During training, the model takes the original ink as input, extracts its style, and uses it, together with the original label, to reconstruct the original ink in a teacher-forced manner, minimizing the negative log-likelihood.
At training time our model is given the original ink and original label, but during inference it is given the original ink and the spell-corrected label. To stop the model from memorizing the original ink through the style layer, for each training sample we mask each feature of the style vector with a masking probability, chosen for each sample between 0 and 1. 

During inference our model takes as input the original ink and the spell-corrected label, as well as a \textit{similarity} (\textit{sim}) value, which controls the amount of information that will be allowed to flow through the style  (given a \textit{sim} value of X\%, we mask the last 100-X\% of the features in the style vector).
By increasing the amount of information passed through the style, we increase the similarity of the output to the original ink. As increased similarity reduces the recognizability of the result, the \textit{sim} hyperparameter explicitly controls the metric trade-off described in Section~\ref{subsec:tradeoff}. 

Figure~\ref{fig:comparison} show the effect of the masking of the style vector on the outputs of our model. Note how Inkorrect preserves both global ink-level features like size and cursiveness, and local ones, like the way letters are written and connected, and how similarity gradually decreases as we increase the masking of the style. DW performs comparably to our approach for simple inks that are not cursive, but fails to preserve the connectivity pattern and sometimes size, since it lacks global features. 
Ink-level features do not encode high frequency details, resulting in a more smooth appearance. For the same reason, in the middle example Inkorrect connects "in" while it is disconnected in the original ink, since this is more consistent with the overall cursiveness of writing.  

\paragraph{Implementation details} 

We select the best model as the one with the lowest $CER$ on the validation data when using a \textit{sim} value of 1. Using NLL instead leads to overfitting.
We use Adam~\cite{kingma2014adam} (LR=1e-3, batch size 256). 
We sample with high “bias”~\cite{graves2013generating} to produce more accurate results (after selecting the mixture component from which to sample the next point, we take the mean of the Gaussian for the point, instead of sampling from it).

\textbf{Limitations.} Fixed style dimension does not scale to very long inputs and can be aided by variable length style~\cite{skerry2018towards} or by reinstating character segmentation.



%% file: results.tex
\section{Results}
\label{sec:results}



To ensure the resilience of our approach, we use two datasets that exhibit  a variety of writing styles, input lengths, stroke orders, and come from two different languages. Since they feature only ink-label pairs, we augment them with spell-corrected versions of the labels.

\begin{figure}[!h]
    \centering
    \includegraphics[trim=65 20 65 25,clip,width=0.48\textwidth]{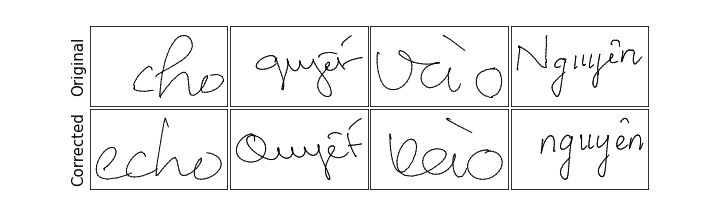}
    \caption{Examples of DISC in \textbf{HANDS-VNOnDB} with \textit{sim}=1.0. Note how correcting "v" to "x" with too much similarity hurts recognizability in the third column.}
    \label{fig:vietnamese}
\end{figure}

\textbf{HANDS-VNOnDB}~\cite{nguyen2018database} (example in Figure~\ref{fig:vietnamese}) contains 66k single Vietnamese words in training set, featuring a lot of cursive samples and delayed strokes for diacritics. 

\textbf{DeepWriting} dataset~\cite{aksan2018deepwriting} consists of approximately 33k training and 700 test inks, most of length around 300 points, containing several words, partly extracted from IAMOnDB~\cite{liwicki2005iam} and partly collected by the authors. 

These two datasets cover two major spelling correction use-cases: while spelling corrections are typically applied to single words (as in \textbf{HANDS-VNOnDB}), changing a single word in a handwritten line may affect other words (ie the corrected word may overlap them if it becomes longer), requiring to either (1) aligning the newly synthesized words with the old ones and reflowing the rest, or (2) resynthesizing the whole line. (1) is a complex error-prone procedure requiring several models or heuristics, so we explore option (2) using the \textbf{DeepWriting} dataset.

\paragraph{Spell-corrected label generation} Spelling correction typically replaces a word with another known (ie dictionary) word. Based on the list of most common English misspellings 
, most spelling corrections (71\%) have a Levenstein edit distance of 1 to the original word, and most of the rest have edit distance of 2. We use this information to generate the spell-corrected labels: for each sample, we pick one word and replace it with a random dictionary word at the edit distance of 1 or 2 with 71/29\% probability. 

\subsection{Human evaluation}
Here we present the results of the human evaluation to answer the questions formulated in Section~\ref{subsec:protocol}. We performed side-by-side comparison with DW method on the \textbf{DeepWriting} test set. The order of spell-corrected images is randomized, and for each sample 3/10 different users saw the sample. 

To understand whether recognizability and similarity of inks correlate with human preference, we use the information about the pairwise comparisons to determine the degree to which the proposed recognizability/similarity metrics follow human judgment below. Due to their relative ease, pairwise comparisons are used extensively in human evaluations~\cite{lau2014machine}.

\paragraph{Criteria for preferring a particular spelling correction} 
In the open-ended feedback, 9/10 participants stated that the first thing they have looked at was the recognizability (can they parse the spelling corrected ink as the intended label), and if the answer was yes for both samples, they start looking at the similarity traits (letter size, angle, cursiveness, shape of individual letters, etc.)

\paragraph{Recognizability-preference correlation}
When one of two inks is correctly recognized by the recognizer, humans prefer it (73\% of the cases, 78\% for recognition in top-10 candidates). 
If the recognized sample is from Inkorrect, it is preferred in 90\% of the cases, and in 62\% of the cases if it is from DW.
This gap can be explained by additional potentially desirable traits, such as a high degree of smoothness. Overall, the study thus shows the relevance of recognizability as an automated evaluation of DISC.

\paragraph{Similarity-preference correlation}
Since we've established that people prefer samples with higher similarity \textbf{iff} they can recognize both inks, all evaluations we performed in this section have been limited to samples where outputs of both models can be recognized by the recognizer.

Since similarity is a continuous, rather than a binary value, and, as any automated metric, is a subject to noise, for slight differences, the amount of noise can overwhelm the signal. We want to show is that as the differences between two models get larger, lower $\mathit{CDE}$ almost always correlates with human preference. Similar evaluation has been done, for example, in~\cite{musat2011improving}.

\begin{figure}
    \centering
    \includegraphics[trim=10 10 10 10,clip,width=0.48\textwidth]{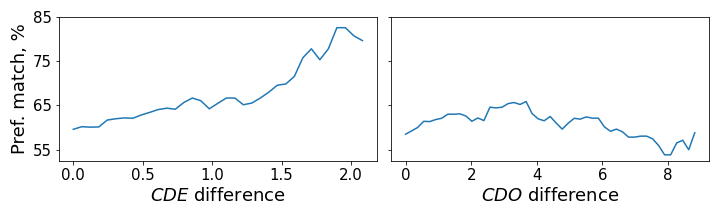}
    \caption{Relationship between the difference in the distance metric and percentage of cases where lower distance matches the human preference. Curves for $\mathit{CDE}$ and $\mathit{CDO}$ are plotted for all values of distance difference for which there are at least 50 samples from the human evaluation with such differences.}
    \label{fig:preference}
\end{figure}

We rank all samples by the threshold of difference in distance metric $|D_\text{Inkcorrect}-D_\text{DW}|$ ($D$ is the similarity metric, $\mathit{CDO}$ or $\mathit{CDE}$, between original and sample synthesized by the two methods) and show that as this threshold grows, the samples with the lower distance are preferred by humans more frequently. As shown by the Figure~\ref{fig:preference}, lower values of $\mathit{CDE}$ agree with human preferences up to 85\%, while it is only up to 66\% for $\mathit{CDO}$. This underscores that $\mathit{CDE}$ is closer to human evaluation than previously used $\mathit{CDO}$.

\paragraph{Comparison to DW} Our model is preferred in $66\%\pm 4.7\%$ of the cases. 
Automated evaluation, comparing samples based on recognizability-then-similarity, agrees with human evaluation in 79.4\% of the cases, further validating agreement between automated metrics and human judgement and providing a way for practitioner to compare two models.

\subsection{Automated evaluation metrics}

\begin{figure}
    \centering
    \includegraphics[trim=11 13 11 11,clip,width=0.48\textwidth]{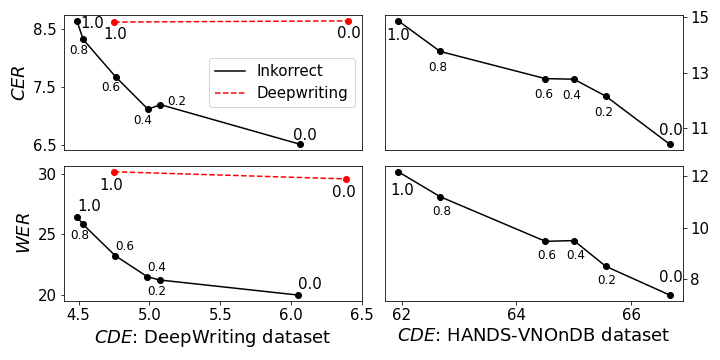}
    \caption{$\mathit{CER}$, $\mathit{WER}$ and $\mathit{CDE}$ values for different versions of Inkorrect and DW. Trade-off between similarity and recognizability is clearly visible. Inkorrect has better metric values at both ends, and for \textit{sim}=0.8, our method has strictly better results on all axes. 
    }
    \label{fig:metrics}
\end{figure}

After validating the recognizability and similarity metrics, in this section we used them study the performance of the proposed method and compare it to the state of the art.
We further show that by varying the $sim$ value, we can trade-off between the similarity ($\mathit{CDE}$) and recognizability ($\mathit{CER}$/$\mathit{WER}$). 

Since DW does not present a way to vary the amount of information passing through the style layer, we report only two points for their method: $sim=1.0$ (extracting the style from the original ink) and $sim=0.0$ (model not conditioned on any input). For our model, we vary the percentage of non-masked features between 1.0 and 0.0 with 0.2 increments.

As visible in Figure~\ref{fig:metrics}, for maximum similarity, Inkorrect has better $\mathit{CDE}$ and $\mathit{WER}$ with approximately similar $\mathit{CER}$. When 20\% of features are masked, Inkorrect is better on all axes than DW conditioned on the original ink, and without any style information, Inkorrect is preferred to the version of DW not conditioned on original ink. We hypothesize that Inkorrect lower error rates are due our model not being conditioned on potentially incorrect segmentation information. 

\subsection{Metric Tradeoff}
\label{subsec:ablation}
In Inkorrect, the tradeoff between spell-corrected ink similarity and recognizability can be changed by varying the amount of information passing through the style layer. This can be achieved not only during inference time by varying the similarity parameter, but also by: \textbf{(i)} Varying the dimensionality of the style layer and \textbf{(ii)} using a different masking scheme during training.

\begin{table}
\centering
\begin{tabular}{lrr}
\toprule
\textbf{Style dimension}  & $\mathit{CDE}$ & $\mathit{CER}@1$ \\
\midrule
250 & 4.4 & 9.3 \\
50 & 4.5 & 8.6 \\
10 & 4.7 & 8.0 \\
0 & 6.0 & 6.4 \\
\bottomrule
\end{tabular}
\caption{Different style dimension results in different trade-offs between similarity and recognizability in the final model (\textit{sim}=1.0). 
}
\label{tab:dimension}
\end{table}

\begin{table}
\centering
\begin{tabular}{lrr}
\toprule
\textbf{Masking scheme}  & $\mathit{CDE}$ & $\mathit{CER}@1$ \\
\midrule
Feature masking & 4.5 & 8.6 \\
Timestep skipping (continuous) & 4.6 & 8.5 \\
Timestep skipping (random) & 4.7 & 8.0 \\
No masking & 4.3 & 12.9 \\
\bottomrule
\end{tabular}
\caption{Different ways of masking information passing through the style layer. Feature masking refers to our approach outlined in Section~\ref{sec:method}. Timestep masking refers to skipping continuous or random parts on the input sequence in attempt to teach the model that ink passing through the style layer does not perfectly match spell-corrected label. Last row is for absence of any masking scheme.}
\label{tab:masking}
\end{table}

As shown in Table~\ref{tab:dimension}, having larger style dimension allows to reach lower $\mathit{CDE}$ at the cost of higher $\mathit{CER}$ and vice versa. For human evaluation, we decided to use the model with dimension 50, because unlike the models with dimensions of 10 and 250, it achieves both similarity and recognizability strictly better than those of DW.

Table~\ref{tab:masking} shows comparison of our approach of masking feature dimensions to 3 alternatives. As mentioned in Section~\ref{sec:method}, absence of any masking scheme results in a model that learns to do a very good reconstruction, but fails to generalize to unseen labels. Approaches to masking timesteps instead of ink features lead to lower similarity. We attribute this to the fact that it is much harder to learn to match the parts of the ink that are not masked to the whole ink that needs to be written (this could potentially be aided by finding an alignment between the points of the original ink and the letters of spell-corrected label, either by having access to character segmentation or via an attention mechanism). 

The masking result in particular shows the importance of having fine-grained control over the recognizability-similarity tradeoff, as significant gains can be made on one metric with little impact on the other.